\pdfoutput=1

\documentclass[11pt]{article}

\usepackage[final]{acl}
\usepackage{sectsty}
\sectionfont{\fontsize{12}{8}\selectfont}

\usepackage{times}
\usepackage{latexsym}
\usepackage{xcolor}

\usepackage[T1]{fontenc}

\usepackage[utf8]{inputenc}

\usepackage{microtype}

\usepackage{inconsolata}

\fontsize{15}{15}\title{Maha Bhaashya at SemEval-2024 Task 6: Zero-Shot Multi-task Hallucination Detection}

\author{
    \fontsize{12}{14}\bfseries Patanjali Bhamidipati$^\dagger$\\
    \fontsize{12}{14} IIIT Hyderabad \\
    \texttt{patanjali.b@research.iiit.ac.in}
    \ \ \ \ \ \ \ \
    \\\And
    \fontsize{12}{14}\bfseries Advaith Malladi$^\dagger$\\
    \fontsize{12}{14} IIIT Hyderabad \\
    \texttt{advaith.malladi@research.iiit.ac.in}
    \AND
    \fontsize{12}{14}\bfseries Manish Shrivastava\\
    \fontsize{12}{14} IIIT Hyderabad \\
    \texttt{m.shrivastava@iiit.ac.in}
    \ \ \ \ \ \ \ \
    \\\And
    \fontsize{12}{14}\bfseries Radhika Mamidi\\
    \fontsize{12}{14} IIIT Hyderabad \\
    \texttt{radhika.mamidi@iiit.ac.in}
}

\begin{document}
\maketitle
\fontsize{12}{14}\begin{abstract}
\fontsize{10}{14}In recent studies, the extensive utilization of large language models has underscored the importance of robust evaluation methodologies for assessing text generation quality and relevance to specific tasks. This has revealed a prevalent issue known as hallucination, an emergent condition in the model where generated text lacks faithfulness to the source and deviates from the evaluation criteria. In this study, we formally define hallucination and propose a framework for its quantitative detection in a zero-shot setting, leveraging our definition and the assumption that model outputs entail task and sample specific inputs. In detecting hallucinations, our solution achieves an accuracy of 0.78 in a model-aware setting and 0.61 in a model-agnostic setting. Notably, our solution maintains computational efficiency, requiring far less computational resources than other SOTA approaches, aligning with the trend towards lightweight and compressed models.
\end{abstract}
\def\thefootnote{$\dagger$}\footnotetext{The authors contributed equally to this work.}

\section{Introduction}
The contemporary landscape of Natural Language Generation (NLG) is marked by a confluence of complexities, wherein two primary challenges emerge as focal points of concern. Firstly, the prevalent neural models within NLG frameworks consistently produce outputs that exhibit linguistic fluency yet suffer from inaccuracies. Secondly, the current evaluation metrics, vital for evaluating the effectiveness of NLG systems, demonstrate a significant inclination towards fluency measures while neglecting to prioritize accuracy. So, this highlights the need to consider the "truthfulness" of the model's output, i.e its alignment with the source to ensure a comprehensive assessment.\cite{dale2022detecting}

In the realm of NLG applications, the criticality of output accuracy cannot be overstated. A divergence between the fluency and factual correctness of generated content not only undermines the utility of NLG systems but also engenders substantial risks across various domains. Consider, for instance, the domain of machine translation,the production of seemingly plausible yet inaccurate translations not only compromises the integrity of the translated content but also defeats the purpose of facilitating correct translations. 

Likewise, in tasks like definition modeling and paraphrase generation, where accurately conveying semantic meaning is crucial, the presence of incorrect outputs presents notable challenges in upholding the integrity and dependability of the generated content.

\section{SHROOM Dataset}

SHROOM (a Shared-task on Hallucinations and Related Observable Overgeneration Mistakes) dataset is a task-based hallucination detection dataset which is divided into two major categories:

\subsection{Model Aware (MAw)}
Model Aware (MAw) refers to situations where the model under study is known.
\subsection{Model Agnostic (MAg)}
Model Agnostic (MAg) refers to situations where the model under study is not known.\\

The dataset encompasses three major Natural Language Generation tasks, namely: 

\textbf{1. Definition Modeling (DM):}
In this task, models are trained to generate a definition for a given example in context.

\textbf{2. Machine Translation (MT):}
In this task, models aim to generate translations of the given samples.

\textbf{3. Paraphrase Generation (PG):}
In this task, models aim to produce paraphrases of the given text samples. \\

Further, each sample in the train set is populated with information such as task ($task$): indicating what objective the model was trained for, source ($src$): the input passed to the models for the generation, target ($tgt$): the intended reference "gold" standard text that the model ought to generate, hypothesis ($hyp$): the actual model production,also the model-aware dataset is populated with  model name ($model$) used for the task, with the val set additionally being populated with majority-based gold-label ($label$), based on the annotator labels along with the probability values of the sample being hallucination ($p(Hallucination)$) based on the proportion of annotators who claim that the sample is an hallucination.


\section{Definitions}

As described earlier, the SHROOM shared task encompasses of three different taks, Definition Modelling (DM), Paraphrase Generation (PG), and Machine Translation. We define the \textbf{Hallucination} in the context of the specific task at hand. Defining hallucinations individually in the context of a specific task enables detecting hallucinations quantitatively and qualitatively. We offer distinct definitions of hallucinations and methodologies for detecting hallucinations within the context of each of the aforementioned task.

In the context of definition modelling, the model is expected to generate the definition of a word which has been used in the provided context by making use of distributional semantics. Definition modelling models such as flan-t5-definition-en-base \cite{giulianelli-etal-2023-interpretable} are not fully capable of making use of distributional semantics to define a word as used in a context. In a sample where the word W has been used in a setting contrasting to the definition the model has learnt during its training process, the models fails to provide a contextual definition of the word W. Examples for the same have been demonstrated in Table 1 and Table 2. We observe the model outputs a definition of word W which is very similar what it has learnt during its training process. Based on this observation, we assume that the targets provided in the SHROOM dataset have been extracted from the training data of the definition modelling dataset. With this assumption, we define
\textbf{"Hallucination to be an instance where the output generated by the definition modelling model \textit{does not} entail the target output."}
Thereby reducing the hallucination detection task to a Natural Language Inference task in the context of definition modelling.

In the context of paraphrase generation and machine translation, the model's inputs and outputs are anticipated to exhibit semantic equivalence. If the generated paraphrase or translation diverges from semantic equivalence with the source text, they are deemed imperfect paraphrases or translations. Therefore, in the context of paraphrase generation and machine translation, we define \textbf{"Hallucination to be an instance where the paraphrase or translation generated by the model is \textit{not semantically equivalent} to the source."
} This reduces the hallucination detection in the context of given tasks to a semantic equivalence detection task, which could also be framed as bidirectional entailment detection, a variation of the Natural Language Inference task.

The aforementioned definitions of hallucination allow us to simplify the hallucination detection task to a Natural Language Inference task, thereby enabling us to qualitatively and quantitatively detect hallucinations.

In a more generic setting, we provide a definition of hallucinations that can be adapted to any task to effectively detect them. We define \textbf{"hallucinations as instances where the output generated by the model is \textit{not faithful} to the input or the training data of the model. If the model generates information that is contradictory to the model's training data or the input to the model, it can be termed as a hallucination."}

\begin{table*}
    \centering
    \begin{tabular}{|p{0.9\linewidth}|}
        \hline
        \textbf{Example 1: Definition Modeling (DM)} \\
        \hline
        \textbf{Model Input:} I went into the \textit{water bottle} to withdraw cash. What is the definition of \textit{water bottle}? \\
        \textbf{Model Output:} A container for holding liquids. \\
        \textbf{Expected Output:} A financial institution such as a bank or ATM to withdraw cash \\
        \textbf{Model:} flan-t5-definition-en-base \\
        \hline
    \end{tabular}
    \caption{\label{citation-guide}
A Table showing distinction between the model output and the expected output where the model fails to understand the contextual definition of the term \textit{water bottle}
}
    
    \vspace{0.5cm} 
    
    \end{table*}
 \begin{table*}
    \centering
    \begin{tabular}{|p{0.9\linewidth}|}
        \hline
        \textbf{Example 2: Definition Modeling (DM)} \\
        \hline I jumped into the flaxcron to do some swimming. What is the meaning of \textit{flaxcron}? \\
        \textbf{Model Output:} A slender, slender \\
        \textbf{Expected Output:} A pool of water. \\
        \textbf{Model:} flan-t5-definition-en-base \\
        \hline
    \end{tabular}
    \caption{\label{citation-guide}
A Table showing distinction between the model output and the expected output where the model fails to understand the contextual definition of the term \textit{flaxcron}
}
    
    \vspace{0.5cm} 
    
    \end{table*}   

\section{Methodology}
Grounding to the above definitions, the experimental setup  we designed goes on to quantify the alignment of the model's output ($hyp$) with either the source ($src$) or the target ($tgt$) based on the task ($task$) the data sample corresponds to.

We propose that examining the \textbf{entailment} relationship between the model's output ($hyp$) and either the source ($src$) or the target ($tgt$) (which is also inherently linked to the source ($src$)), depending on the task, sheds light on data samples that are \textbf{not} "detached" from the source. Consistent with our initial hypothesis that hallucinations occur when samples are "detached" from the source, this approach based on Natural Language Inference (NLI) can effectively aid in hallucination classification.
\begin{itemize}
  \item In the context of definition modelling, \textbf{if the $hyp$ does not entail the $tgt$}, the sample has been classified as \textbf{Hallucination}.
  \item In the context of machine translation and paraphrase generation, we check equivalence through bidirectional entailment. \textbf{If the $hyp$ does not bidirectionally entail the $src$}, the sample has been classified as \textbf{Hallucination}
  \item In the context of machine translation and paraphrase generation, we verify our hypothesis of semantic equivalence between the $src$ and $hyp$ by comparing the performance metrics in the case of both unidirectional and bidirectional entailment.
\end{itemize}
    Recent research heavily relies on large language models (LLMs) to benchmark various natural language understanding and generation (NLG) tasks. However, this practice extends to hallucination detection as well, which we find ironic and counterproductive, considering LLMs' inherent tendency to hallucinate. Using LLMs for hallucination detection presents two major drawbacks. Firstly, their computational demands are significant, making them an expensive solution \cite{bai2024efficiency}. Secondly, the lack of complete interpretability in LLMs renders them unreliable for this task \cite{singh2024rethinking}.

\section{Results}

After several experiments with the above methodology, leveraging the accuracy and the Spearman correlation ($\rho$) metrics, we have bench-marked the hallucination detection task on the SHROOM validation and test sets to achieve an accuracy of 0.78 in model-aware and 0.61 on model-agnostic test sets respectively. For our analysis let us take only the accuracy metric into account.\\
The bench-marking saw a utilisation of open-source pre-trained Natural Language Inference (NLI) models available on Hugging Face. Several experiments brought out interesting observations which are worthy discussing. \\ \\
We evaluated the following models:
\begin{enumerate}
    \item MoritzLaurer/DeBERTa-v3-base-mnli-fever-anli \textbf{(DeBERTa-1)} \cite{DBLP:journals/corr/abs-2006-03654}
    \item MoritzLaurer/mDeBERTa-v3-base-xnli-multilingual-nli-2mil7 \textbf{(DeBERTa-2)} \cite{DBLP:journals/corr/abs-2006-03654}
    \item ynie/bart-large-snli\_mnli\_fever\_anli\_R1\_R2\_R3-nli \textbf{(BART-1)} \cite{DBLP:journals/corr/abs-1910-13461}
    \item ynie/roberta-large-snli\_mnli\_fever\_anli\_R1\_R2\_R3-nli \textbf{(RoBERTa-1)} \cite{nie-etal-2020-adversarial}
\end{enumerate}

\subsection{Definition Modelling}
For the task of definition modelling, our approach achieves a peak accuracy of 0.748663 using the DeBERTa-2 model in a model-agnostic setting and a peak accuracy of 0.755319 using the RoBERTa-1 model in a model-aware setting. These results are in accordance with our hypothesis that when the model hallucinates, it does not entail the target.

\begin{table}
\centering
\begin{tabular}{lll}
\hline
\textbf{Model } & \textbf{Unidirectional} & \textbf{Bidirectional}\\
\hline
DeBERTa - 1 & 0.783567& 0.755511 \\
DeBERTa - 2 & 0.765531 & 0.717435\\
BART - 1 & 0.769539 & 0.733467 \\
RoBERTa - 1 & 0.757515  & 0.727455 \\
\hline
\end{tabular}
\caption{\label{citation-guide}
Model-agnostic evaluation on (Uni vs Bi) directional entailment.
}
\end{table}
\begin{table}
\centering
\begin{tabular}{lll}
\hline
\textbf{Model } & \textbf{Unidirectional} & \textbf{Bidirectional}\\
\hline
DeBERTa - 1 &  0.596806& 0.570859 \\
DeBERTa - 2 &  0.576846 & 0.586826\\
BART - 1 & 0.610778 & 0.568862 \\
RoBERTa - 1 & 0.612774  & 0.584830 \\
\hline
\end{tabular}
\caption{\label{citation-guide}
Model-aware evaluation on (Uni vs Bi) directional entailment.
}
\end{table}
\begin{table}
\centering
\begin{tabular}{lll}
\hline
\textbf{Model } & \textbf{Unidirectional} & \textbf{Bidirectional}\\
\hline
DeBERTa - 1 & 0.728 & 0.752 \\
DeBERTa - 2 & 0.624 & 0.68\\
BART - 1 & 0.696 & 0.72 \\
RoBERTa - 1 & 0.712  & 0.752 \\
\hline
\end{tabular}
\caption{\label{citation-guide}
Accuracy validation on \textbf{PG} task
}
\end{table}

\begin{table}
\centering
\begin{tabular}{lll}
\hline
\textbf{Model } & \textbf{Unidirectional} & \textbf{Bidirectional}\\
\hline
DeBERTa - 1 & - & - \\
DeBERTa - 2 & 0.722 & 0.754\\
BART - 1 & - & - \\
RoBERTa - 1 & -  & - \\
\hline
\end{tabular}
\caption{\label{citation-guide}
Accuracy validation on \textbf{MT} task.
}
\end{table}
\begin{table}
\centering
\begin{tabular}{llll}
\hline
\textbf{Model } & \textbf{DM} & \textbf{MT} & \textbf{PG}\\
\hline
DeBERTa - 1 & 0.721925 & \textcolor{red}{0.855615} & \textcolor{red}{0.768} \\
DeBERTa - 2 & \textcolor{red}{0.748663} & 0.823529 & 0.704  \\
BART - 1 & \textcolor{red}{0.748663}  & 0.844920 & 0.688 \\
RoBERTa - 1 & 0.711230 & 0.834224 & 0.712\\
\hline
\end{tabular}
\caption{\label{citation-guide}
Model-agnostic evaluation on individual tasks.
}
\end{table}
\begin{table}
\centering
\begin{tabular}{llll}
\hline
\textbf{Model } & \textbf{DM} & \textbf{MT} & \textbf{PG}\\
\hline
DeBERTa - 1 & 0.744681& {0.707447} & {0.208} \\
DeBERTa - 2 & { 0.670213} & 0.718085 & \textcolor{red}{0.224}  \\
BART - 1 & {0.728723}  & \textcolor{red}{0.760638} & 0.208 \\
RoBERTa - 1 & \textcolor{red}{0.755319} & 0.734042 & 0.216\\
\hline
\end{tabular}
\caption{\label{citation-guide}
Model-aware evaluation on individual tasks.
}
\end{table}

\subsection{Paraphrase Generation and Machine Translation}
For the paraphrase generation and machine translation tasks, the observed results confirm our hypothesis that if the source (src) and hypothesis (hyp) are not semantically equivalent, the hypothesis is a hallucination. In hallucination detection for the paraphrase generation task, we observe that bidirectional entailment outperforms the unidirectional entailment approach for all models. Similar results can also be observed for the machine translation task. This provides evidence that in machine translation and paraphrase generation, hallucinations can be detected by checking for semantic equivalency between the source and hypothesis.

\subsection{Overall Analysis}

In the model-agnostic setting, we achieve a peak accuracy of 0.783567 using the DeBERTa-1 model and a peak accuracy of 0.612774 in a model-aware setting using the RoBERTa-1 model. These scores exhibit satisfactory performance of models pre-trained on the Natural Language Inference task for Hallucination Detection.

\

\section{Conclusion}
Our work makes two significant contributions to the study of hallucinations in language models. First, we provide a concrete definition of the term "hallucination," enabling both qualitative and quantitative study and detection of such phenomena. Second, we offer a computationally efficient approach to detect hallucinations in tasks such as definition modeling, machine translation, and paraphrase generation. We frame the hallucination detection task as a function of the input to the generation model and the data used to train it. Our definitions and approaches also provide a framework that can be utilized for hallucination detection in various Natural Language Generation tasks across the spectrum.

\bibliography{custom}

\appendix

\end{document}